\documentclass[11pt]{article}

\usepackage[preprint]{acl}

\usepackage{times}
\usepackage{latexsym}

\usepackage[T1]{fontenc}

\usepackage[utf8]{inputenc}

\usepackage{microtype}

\usepackage{inconsolata}

\usepackage{kotex}
\usepackage{graphicx}
\usepackage{amsmath}
\usepackage{amssymb}
\usepackage{xcolor}
\usepackage{graphicx} 
\usepackage{algorithm}
\usepackage{algcompatible}
\usepackage{booktabs}
\usepackage{multirow}
\usepackage{multicol}
\usepackage{array}
\usepackage{tabularx}
\usepackage{soul}
\usepackage{xspace}
\usepackage{makecell}
\usepackage{pifont} 
\usepackage{enumitem}

\newcommand{\reuni}{%
  \ooalign{\color{black}$\blacklozenge$\cr\hss\textcolor{white}{\scalebox{0.6}{\textbf{?}}}\hss}%
}
\newcommand{\tok}{Thunder-Tok\xspace}
\newcommand{\xmark}{\ding{55}}

\title{Less Is More: Reducing Token Counts \\ Without Compromising Performance
}

\author{
Gyeongje Cho $^{1}$ \quad
Yeonkyoung So $^{1}$ \quad
Sangmin Lee $^{2}$ \quad
Jaejin Lee$^{1,2}$ \\[0.5em]
$^{1}$Graduate School of Data Science, Seoul National University \\
$^{2}$Department of Computer Science, Seoul National University \\[0.5em]
{\small\texttt{gyeongje@aces.snu.ac.kr, kathy1028@snu.ac.kr, sangmin.lee@snu.ac.kr, jaejin@snu.ac.kr}}
}

\begin{document}
\maketitle
\begin{abstract}
Tokenization directly affects the inference efficiency of large language models, since fragmented tokenization increases sequence length and generation cost. Although longer, multi-word tokens can reduce fertility, naively adding them often degrades language model performance. We propose \tok, a subword tokenizer that reduces fertility while preserving downstream performance. \tok first constructs a large seed vocabulary from corpus substrings and filters structurally incomplete candidates, including invalid Unicode byte fragments and word-boundary violations. It then prunes the seed vocabulary using a likelihood-based token score derived from a uniform Jensen lower bound of the training-data probability. Experiments show that \tok reduces fertility by approximately 25\% in English and 9\% in Korean compared with the standard BPE tokenizer while maintaining competitive performance.
\end{abstract}
\section{Introduction}
Chatbot services powered by large language models (LLMs)~\cite{bai2023qwen, grattafiori2024llama, team2024gemma, guo2025deepseek} have become increasingly prevalent and are influencing users' daily activities. As these services are introduced into more diverse and demanding contexts, enhancing model performance remains a primary goal. However, many strategies for improving performance, such as increasing model size or incorporating additional test-time computation, result in higher inference costs~\cite{liu2024deepseek, meta2025llama4}. Consequently, achieving efficient inference is a critical challenge for the practical deployment of LLM-based systems.

Tokenization has received increasing attention as a factor that directly influences inference efficiency~\cite{petrov2023language}. A tokenizer transforms raw text into a sequence of discrete tokens suitable for processing by a language model. Since LLMs process and generate text token by token, the number of tokens generated by the tokenizer determines the number of token-level operations required during inference. Consequently, even if the underlying model remains unchanged, a tokenizer that encodes the same text with fewer tokens can reduce inference computational cost.

\begin{figure}
    \centering
    \includegraphics[width=1.0\linewidth]{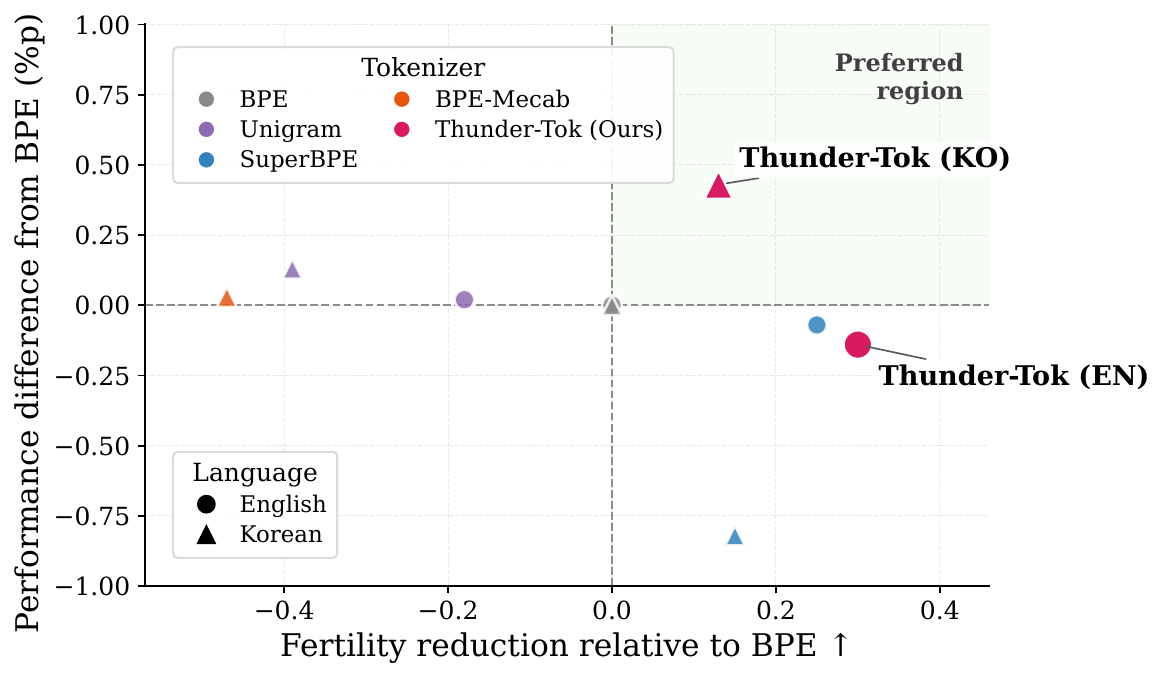}
    \caption{Relative performance-fertility trade-off across tokenizers. \tok moves toward lower fertility while maintaining competitive downstream performance compared with BPE.}
    \label{fig:intro_tradeoff}
\end{figure}

This efficiency is typically measured by \textit{fertility}, which refers to the number of tokens required to represent each word. Lower fertility results in more compact segmentation and reduces the number of token-level operations necessary for processing or generation. Consequently, minimizing fertility is a key tokenizer-level objective for enhancing inference efficiency.

One approach to lowering fertility involves expanding the vocabulary with longer tokens, including those that span multiple words, such as "as well as" or "New York." These expressions frequently function as coherent lexical or semantic units, and representing them as single tokens can reduce fragmentation and shorten tokenized sequences. However, prior vocabulary expansion methods that introduce longer or multi-word tokens have often resulted in decreased language model performance~\cite{kumar2022bpe, schmidt2024tokenization, goldman2024unpacking}. This outcome indicates that simply adding longer tokens is insufficient; the tokenizer must carefully select tokens that enhance compactness without disrupting the statistical structure learned by the language model.

SuperBPE~\cite{liu2025superbpe} addresses this challenge by initially learning a base vocabulary using Byte Pair Encoding (BPE) and subsequently augmenting it with multi-word tokens derived from the learned BPE vocabulary. This strategy enables SuperBPE to reduce fertility without compromising performance. Nevertheless, since this method is fundamentally based on BPE, it does not generalize effectively to non-English languages such as Korean, where linguistic structure and character representation differ substantially~\cite{vemula2025rethinking}.

This paper introduces a new tokenizer, \textit{\tok}, which is designed to enhance model performance while reducing token fragmentation, as measured by fertility. The proposed approach is based on two main principles. First, {\tok} constructs a large seed vocabulary from corpus substrings, filtering out structurally incomplete candidates such as invalid Unicode byte fragments and tokens that violate word boundaries. Second, it prunes this vocabulary using a token score derived from a uniform Jensen lower bound of the training-data probability. This score quantifies the decrease in the lower bound resulting from the removal of each token, allowing {\tok} to iteratively eliminate low-scoring candidates and produce a compact, fixed-size vocabulary. The resulting vocabulary is then employed with likelihood-based segmentation during inference.

Compared to BPE, {\tok} reduces fertility by approximately 25\% for English and 9\% for Korean, while maintaining language model performance. Figure~\ref{fig:intro_tradeoff} illustrates this performance-fertility trade-off relative to BPE: moving right indicates greater fertility reduction, while moving upward indicates better downstream performance. {\tok} moves toward the preferred region in both English and Korean, showing that it reduces token fragmentation without substantially compromising model quality.

The main contributions of this paper are summarized as follows:
\begin{itemize}[leftmargin=*, itemsep=0pt, topsep=2pt]
\item We propose a structurally constrained seed vocabulary construction method that extracts substring candidates from the training corpus and filters out incomplete tokens at both the Unicode-byte and word-boundary levels.
\item We introduce a token scoring strategy based on the reduction in a uniform Jensen lower bound, decomposing each token score into a branching-entropy term and a token-level log-likelihood term.
\item We integrate the proposed vocabulary construction and pruning methods into a new subword tokenizer, {\tok}, which achieves lower fertility and maintains competitive performance in both English and Korean language modeling tasks.
\end{itemize}

\section{Related Work}
Subword tokenization represents the predominant approach in contemporary language models. Prominent methods include Byte Pair Encoding (BPE)~\cite{sennrich2015neural}, the Unigram tokenizer~\cite{kudo2018subword}, and WordPiece~\cite{schuster2012japanese}. The widespread adoption of subword tokenization is largely due to its language-agnostic properties. This approach addresses the challenge of rare or unseen words by decomposing them into smaller, more frequent units, thereby maintaining a manageable vocabulary size.

Extensive prior research has focused on improving tokenizer quality to enhance language model performance. Many of these methods incorporate linguistic structure or domain-specific information. For example, GraphBPE~\cite{shen2024graphbpe} adapts BPE to operate at the grapheme level, while ADAPT-BPE~\cite{balde2024adaptive} expands the vocabulary by introducing domain-specific tokens. Multiple studies~\cite{park2020empirical, kim2021changes, fujii2023different} have shown that integrating subword tokenizers with morphological analyzers improves performance for agglutinative languages such as Korean and Japanese by aligning token boundaries with morphological structures.

Recent research has increasingly focused on reducing token count to improve inference efficiency. One approach involves adding arbitrarily long strings to the vocabulary. However, previous studies~\cite{kumar2022bpe, dagan2024getting, schmidt2024tokenization} demonstrate that indiscriminate inclusion of such tokens can substantially degrade model performance. Conversely, restricting the types of additional tokens can reduce the token count while maintaining or improving performance. For instance, T-MuFin~\cite{lambruschini2023reducing} and VEGAD~\cite{liu2024gold} introduce domain-specific tokens, whereas other methods~\cite{otani2020pre, gee2023multi} incorporate multi-word tokens to enhance outcomes in tasks such as machine translation and text classification. Despite their effectiveness, these approaches often rely on domain expertise or are tailored to specific tasks, limiting their generalizability.

SuperBPE~\cite{liu2025superbpe} seeks to overcome the limitations of BPE-based vocabularies by incorporating multi-word tokens derived from the initial BPE vocabulary. The method constructs a seed vocabulary using standard BPE, followed by an additional BPE-style merging process applied to the resulting tokens. Although this approach effectively reduces fertility in English, it remains fundamentally dependent on BPE. Furthermore, studies~\cite{mager2022bpe, vemula2025rethinking} suggest that BPE-based tokenization is less effective for agglutinative languages, which possess morphological structures distinct from English. As a result, an LLM with the SuperBPE tokenizer may not generalize effectively to agglutinative languages. Experimental results corroborate this observation, as SuperBPE does not outperform standard tokenizers when applied to Korean. These findings highlight the need for alternative tokenization strategies that are both fertility-aware and linguistically robust across diverse languages.

\begin{figure*}[!t]
	\centering
	\includegraphics[width=1.0\textwidth]{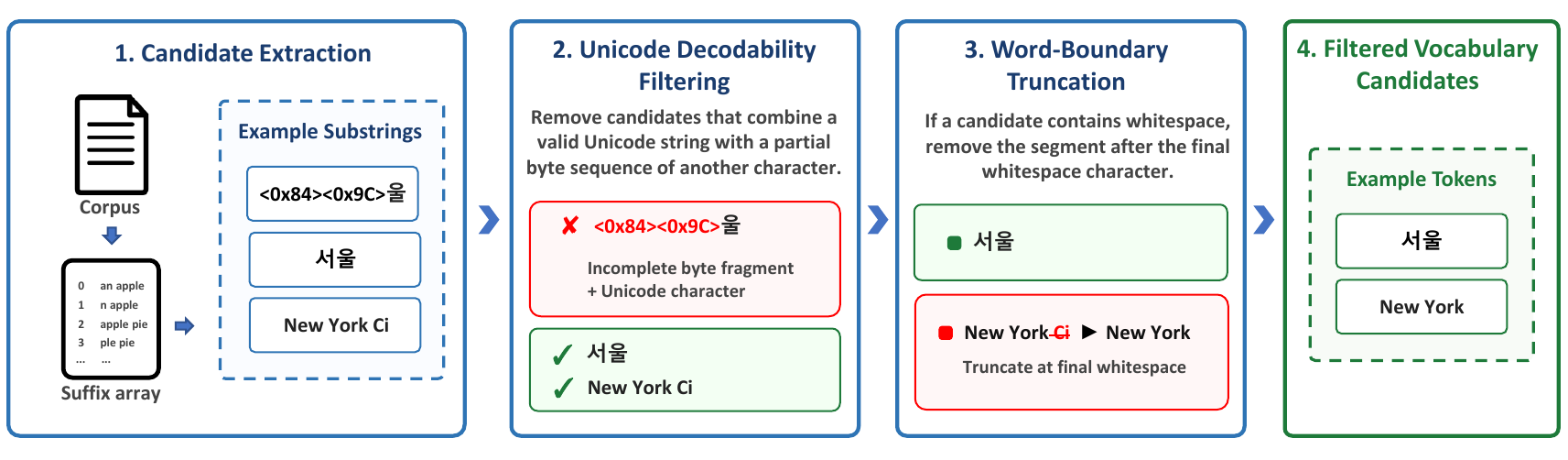}
	\caption{The proposed seed vocabulary construction pipeline begins with substring candidates extracted using a suffix array. The procedure eliminates candidates containing invalid Unicode byte fragments and truncates multi-word candidates that terminate with incomplete trailing word fragments. The filtered candidates constitute the seed vocabulary for subsequent pruning.}
    \label{fig:tokenization}
\end{figure*}

\section{Methods}
\label{sec:methods}
{\tok} is designed to reduce fertility while maintaining language model performance. Similar to SuperBPE~\cite{liu2025superbpe}, it enables tokens to represent sequences longer than a single word, which reduces unnecessary token fragmentation. This section outlines the seed vocabulary construction, vocabulary pruning, and inference procedure of {\tok}.

\subsection{Seed Vocabulary Construction}
\label{sec:init_vocab}
Incorporating longer, high-frequency strings into a tokenizer vocabulary can reduce the average number of tokens required to represent text. However, adding arbitrary strings solely to minimize token count may negatively impact the performance of downstream language models~\cite{kumar2022bpe, schmidt2024tokenization, dagan2024getting}.

Previous research demonstrates that structurally incomplete tokens can be detrimental at both the byte and linguistic boundary levels. Specifically, \citet{jang2025improbable} demonstrate that undecodable byte-level substrings can degrade language model performance by introducing structurally invalid tokens. For example, in the Korean word "서울" (Seoul), the character "서" is encoded in UTF-8 as "<0xEC><0x84><0x9C>". A substring such as "<0x84><0x9C>울" combines an incomplete byte fragment of "서" with the complete character "울."
\citet{pawar-etal-2025-broken} further demonstrate that this issue extends beyond byte-level validity and can also occur at the word level.
For instance, a token such as "New Yo" extracted from "New York" is a valid Unicode string. However, it merges a complete word with only a fragment of the following word, resulting in a structurally incomplete multi-word token.

To address this issue, we explicitly define incomplete tokens within the context of vocabulary construction and introduce a filtering procedure to eliminate them from the seed vocabulary. In this work, an incomplete token refers not only to an undecodable byte sequence but also to any Unicode-decodable string that violates natural word boundaries by attaching a partial word fragment to a complete word.

We begin with the corpus's suffix array, from which substring candidates are extracted. Because these candidates are generated as arbitrary byte-level substrings, some may contain incomplete Unicode byte sequences. We therefore first apply Unicode decodability filtering. A candidate is removed if it combines a valid Unicode string with a partial byte sequence of another character. In contrast, isolated partial bytes corresponding to a single Unicode character are retained at this stage. For example, a candidate such as "<0x84><0x9C>울" is excluded because it combines an incomplete byte fragment with a Unicode-decoded character. The isolated fragment "<0x84><0x9C>" is retained at this stage, since it consists only of partial bytes from a single Unicode character "서."

After Unicode decodability filtering, we further apply word-boundary truncation. Even if a candidate is Unicode-decodable, it may still end in the middle of a word because it is extracted as an arbitrary substring of the corpus. To address this, when a candidate contains one or more whitespace characters, we truncate it at the final whitespace character by removing the segment that follows. This process removes incomplete trailing word fragments while retaining the portion more likely to align with complete word boundaries. For example, from the phrase "New York," a candidate such as "New Yo" contains a whitespace character but ends with the incomplete fragment "Yo." In this case, we truncate the candidate at the final whitespace character, yielding "New." Additional implementation details of the proposed procedure are provided in Appendix~\ref{app:initial_vocab_construction_details}.

\subsection{Token Scoring}
\label{sec:token_scoring}

\begin{algorithm}[t]
   \caption{Vocabulary Construction}
   \label{alg:vocabulary_construction}
   \begin{small}
\hspace*{\algorithmicindent} \textbf{Input:} a set of sentences $Y$\\
\hspace*{\algorithmicindent} \textbf{Input:} a vocabulary size $S$\\
\hspace*{\algorithmicindent} \textbf{Input:} a removal threshold percentage $m$ \\
\hspace*{\algorithmicindent} \textbf{Output:} a vocabulary $X$ of size $S$\\
\begin{algorithmic}
    \STATE $X = \bigcup_{y \in Y} \{\, x \mid x \text{ is a substring of } y$
    \STATE \hspace{7.2em} $\text{\& sequence of single unit type} \,\}$
   \REPEAT
   \STATE Calculate $\Delta BE(x)$ and $\Delta \ell(x)$
   \FOR{each $x \in X$}
   \STATE $score[x]$ = $\Delta \ell (x) - \Delta BE(x)$
   \ENDFOR
   \STATE $L = $ a sorted list of $x \in X$ in the increasing
   \STATE {\hskip3.0em}  order of $score[x]$
   \STATE $X = X - \{\mbox{the first} \; m\% \;\mbox{tokens in} \; L \}$ 
   \UNTIL $|X| = S$ 
\end{algorithmic}
\end{small}
\end{algorithm}

The previous section extracted a large set of candidate tokens for the vocabulary. This section describes the procedure for selecting tokens to construct a final vocabulary of a specified size.

In language model training, the objective is to construct a tokenizer that enables the model to assign high probability to the training dataset $D$, that is, to maximize $P(D)$ under the model. Because directly maximizing $P(D)$ is computationally challenging, this objective is approximated by maximizing a uniform Jensen lower bound of the dataset probability.

Within this framework, the importance of each candidate token is quantified by the decrease in the uniform Jensen lower bound~\cite{jensen1906fonctions} resulting from its removal. Intuitively, this bound replaces the original log-sum objective with the average of token-level log contributions, which provides a tractable surrogate for the dataset log-likelihood. Calculating the exact change in the bound requires re-estimating the complete tokenization statistics after each possible removal, which is computationally intensive. Consequently, a local approximation is used that updates only the contributions of the removed token and its replacement sequence. Tokens whose removal results in a small decrease in the lower bound are considered less important, whereas those whose removal leads to a large decrease are deemed important. Tokens with low scores are iteratively pruned until the vocabulary reaches the target size. The overall procedure is summarized in Algorithm~\ref{alg:vocabulary_construction}.

Starting from the decomposition in
Appendix~\ref{app:token_scroing_derivation}, the token-dependent part of the uniform Jensen lower bound can be written as,

\begin{small}
\begin{equation}
\label{eq:jensen_lower_bound_main}
\mathcal{J}(X)
=
\sum_{x' \in X} f(x')\log P(x')
-
\sum_{x' \in X} f(x')BE(x')
\end{equation}
\end{small}
where
\begin{small}
\begin{equation*}
BE(x)
=
-\sum_{y \in Y_x} P(y \mid x)\log P(y \mid x).
\end{equation*}
\end{small}

In this context, $BE(x)$ denotes the branching entropy~\cite{jin2006unsupervised} of token $x$, and $Y_x$ represents the set of text chunks containing $x$. The first term in Equation~\ref{eq:jensen_lower_bound_main} corresponds to the token-level log-likelihood under a unigram assumption, whereas the second term quantifies the uncertainty over possible text chunks conditioned on each token.

Let $Z_x=(z_1,\ldots,z_m)$ be the token sequence obtained by re-tokenizing the string represented by $x$ after its removal from the vocabulary. The score of $x$ is defined as the approximate decrease in $\mathcal{J}(X)$ resulting from the removal of $x$:

\begin{small}
\begin{equation}
\label{eq:token_score_expanded}
\begin{split}
Score(x)
\approx
\mathcal{J}(X)-\mathcal{J}(X\setminus\{x\})\\
= \Delta \ell(x) - \Delta BE(x),
\end{split}
\end{equation}
\end{small}
where
\begin{small}
\begin{equation*}
\label{eq:delta_ell}
\begin{split}
\Delta \ell(x) =
f(x)
\left(
\log P(x)
-
\sum_{z \in Z_x}\log P'(z)
\right)
\end{split}
\end{equation*}
\end{small}
and
\begin{small}
\begin{equation*}
\label{eq:delta_be}
\begin{split}
\begin{split}
\Delta BE(x)
= f(x)BE(x)
& +
\sum_{z \in Z_x} f(z)BE(z)\\
& -
\sum_{z \in Z_x} f'(z)BE'(z).
\end{split}
\end{split}
\end{equation*}
\end{small}
This formulation clarifies the scoring criterion, where primed quantities are computed under the pruned vocabulary.
The first term, $\Delta \ell(x)$, quantifies the decrease in token-level likelihood when occurrences of $x$ are replaced by the alternative token sequence $Z_x$. The second term, $\Delta BE(x)$, quantifies the change in branching-entropy contribution after occurrences of $x$ are redistributed to the replacement tokens. A larger score indicates that removing $x$ is estimated to cause a greater decrease in the lower-bound objective, making $x$ more important to retain. Conversely, tokens with small approximate scores can be pruned with relatively low estimated impact. Detailed definitions of $\Delta \ell(x)$ and $\Delta BE(x)$ are provided in Appendix~\ref{app:token_score_computation}.

\subsection{Tokenization Method}
After constructing a fixed-size vocabulary, {\tok} tokenizes new input sentences using a likelihood-based segmentation strategy. Consistent with the Unigram Tokenizer~\cite{kudo2018subword}, it selects the token sequence that maximizes the sum of token log probabilities.

This inference procedure aligns with the independence assumption applied during vocabulary construction, in which segmentation likelihood is estimated from token-level probabilities. While {\tok} employs branching entropy to evaluate candidate tokens during vocabulary construction, this metric is not used during inference. Branching entropy measures the contextual flexibility of a token within the training corpus, making it valuable for vocabulary selection; however, it does not indicate the probability of a token in a specific input sentence.

Consequently, after the vocabulary is finalized, {\tok} conducts inference based solely on token likelihood, thereby preventing the application of corpus-level vocabulary selection criteria to sentence-level segmentation.

\begin{table*}[ht]
\centering
\small

\setlength{\tabcolsep}{3pt} 
\resizebox{0.7\linewidth}{!}{
\begin{tabular}{llccccccc}
\toprule
\multirow{2}{*}{Size} & \multirow{2}{*}{Tokenizer} & \multicolumn{7}{c}{English Benchmarks (\%)} \\ \cmidrule(l){3-9} 
 &  & ARC-C & HellaSwag & Winogrande & PiQA & CSQA & RACE & Avg (Std) \\ \midrule
\multirow{4}{*}{560M} & BPE      & 28.86 & 38.16 & 51.76 & 65.93 & 33.79 &  30.89 & 41.56 (0.37) \\
                      & Unigram & 29.30 & 38.00 & 52.30 & 65.79 & 33.91 & 30.18 & \underline{41.58} (0.40) \\
                      & SuperBPE & 28.53 & 40.54 & 51.96 & 66.62 & 33.84 & 30.22 & \textbf{41.95} (0.40) \\
                      & \tok & 29.03 & 38.63 & 51.89 & 65.53 & 34.18 & 30.09 & 41.56 (0.39) \\ \midrule
\multirow{4}{*}{2.5B}  & BPE      & 41.30 & 50.33 & 56.35 & 71.38 & 41.44 & 34.35 & \underline{49.19} \\
                      & Unigram & 41.39 & 51.58 & 55.72 & 71.11 & 41.61 & 33.78 & \textbf{49.20} \\
                      & SuperBPE & 39.42 & 46.74 & 55.72 & 72.47 & 42.92 & 34.16 & 48.57 \\
                      & \tok & 40.78 & 50.24 & 54.14 & 71.11 & 43.98 & 33.20 & 48.91 \\ \bottomrule
\end{tabular}
}
\vspace{-0.5\baselineskip}
\caption{English Benchmark Performance by Model Size and Tokenizer}
\label{tab:english_benchmarks}
\end{table*}
\begin{table*}[ht]
\centering
\small

\setlength{\tabcolsep}{3pt} 
\resizebox{0.75\linewidth}{!}{
\begin{tabular}{llccccccc}
\toprule
\multirow{3}{*}{Size} & \multirow{3}{*}{Tokenizer} & \multicolumn{7}{c}{Korean Benchmarks (\%)} \\ \cmidrule(l){3-9} 
 &  & Ko & Kobest & Ko & Kobest &  \multirow{2}{*}{CLIcK} & Ko & \multirow{2}{*}{Avg (Std)} \\
 &  & ARC-C & HellaSwag & Winogrande & COPA & & LAMBADA & \\ \midrule
\multirow{5}{*}{560M} & BPE      & 24.92 & 53.72 & 52.15 & 64.88 & 34.08 & 86.23 & 52.66 (0.18) \\
                      & BPE Mecab & 24.13 & 50.00 & 53.76 & 66.12 & 33.64 & 88.50 & 52.69 (0.36) \\
                      & Unigram & 24.88 & 53.16 & 52.60 & 65.30 & 34.29 & 86.48 & \underline{52.79} (0.90) \\
                      & SuperBPE & 25.00 & 46.68 & 51.93 & 64.52 & 36.15 & 86.75 & 51.84 (0.55) \\
                      & \tok & 25.11 & 53.12 & 52.60 & 65.28 & 34.17 & 88.28 & \textbf{53.09} (0.16) \\ \midrule
\multirow{5}{*}{2.5B}  & BPE      & 26.99 & 55.20 & 53.74 & 70.80 & 38.60 & 91.75 & 56.23 \\
                      & BPE Mecab & 28.11 & 57.80 & 53.59 & 72.80 & 38.85 & 92.73 & \underline{57.31} \\
                      & Unigram & 27.68 & 57.20 & 53.12 & 73.50 & 37.84 & 91.44 & 56.80 \\
                      & SuperBPE & 27.51 & 56.00 & 54.85 & 72.50 & 39.00 & 90.51 & 56.73 \\
                      & \tok & 28.88 & 58.20 & 53.75 & 72.30 & 39.50 & 91.93 & \textbf{57.43} \\ \bottomrule
\end{tabular}
}
\vspace{-0.5\baselineskip}
\caption{Korean Benchmark Performance by Model Size and Tokenizer}
\label{tab:korean_benchmarks}
\end{table*}

\section{Experiments}
\label{sec:experiments}

\tok is evaluated from three perspectives: downstream language-model performance, tokenization efficiency, and the contribution of each proposed component. \tok is compared with widely used subword tokenizers, including  BPE~\cite{sennrich2015neural}, the Unigram tokenizer~\cite{kudo2018subword}, and SuperBPE~\cite{liu2025superbpe}. For Korean, BPE-Mecab~\cite{park2020empirical}, which uses a Korean morphological analyzer\footnote{https://github.com/hephaex/mecab-ko} as a language-specific preprocessing step, is also included. All tokenizers are trained with a fixed vocabulary size of 128K. Details of tokenizer configurations, pre-tokenization rules, and tokenizer training data are provided in Appendix~\ref{app:tokenizer_details}.

\subsection{Experimental Setup}
Language models are trained using the LLaMA3 architecture~\cite{grattafiori2024llama}. In accordance with prior work~\cite{li2024datacomp, lotz2025beyond}, which demonstrated that sub-billion-parameter models can serve as effective proxies for tokenizer comparisons, 560M-parameter models are trained for repeated experiments. Additionally, 2.5B-parameter models are trained to assess whether the observed trends persist at a larger scale.

To ensure a fair comparison, all models are trained with an identical model-token budget using the same raw training corpus. Since lower-fertility tokenizers generate fewer tokens per corpus pass, their training streams may repeat the shared corpus more frequently to achieve the fixed token budget. All tokenizers are exposed to the same set of unique raw documents. This experimental design enables comparison of tokenizers under an equivalent number of token-level training updates while controlling for the underlying raw data. Further details regarding model architecture, training hyperparameters, and training-stream construction are provided in Appendix~\ref{sec:experiment_details}.

Evaluation is conducted using both MCQA and generation-oriented benchmarks. The English evaluation includes ARC~\cite{allenai:arc}, HellaSwag~\cite{zellers2019hellaswag}, WinoGrande~\cite{sakaguchi2021winogrande}, PIQA~\cite{bisk2020piqa}, CommonsenseQA~\cite{talmor-etal-2019-commonsenseqa}, RACE~\cite{lai2017race}, IF-Eval~\cite{zhou2023instructionfollowingevaluationlargelanguage}, and GSM8K~\cite{cobbe2021gsm8k}. The Korean evaluation uses Ko-ARC, Ko-WinoGrande, Ko-LAMBADA, Ko-IF-Eval, and Ko-GSM8K~\cite{mcrlkorean2025}, together with KoBEST HellaSwag and KoBEST COPA~\cite{jang2022kobest} and CLIcK~\cite{kim2024click}. Details of the benchmarks and evaluation protocol are provided in Appendix~\ref{sec:benchmarks}.

\subsection{Downstream Performance}
\label{sec:performance}

The effect of tokenizer choice on downstream language model performance is evaluated by comparing tokenizers across English and Korean MCQA benchmarks and generation-oriented benchmarks. English MCQA results are summarized in Table~\ref{tab:english_benchmarks}, and Korean MCQA results are reported in Table~\ref{tab:korean_benchmarks}.

SuperBPE achieves the highest average performance on the English MCQA benchmarks. In the 560M setting, the performance gap between SuperBPE and other tokenizers is minimal. The largest observed difference is 0.39\%p, which is comparable to the standard deviation across repeated runs (approximately 0.4). Consequently, there is no statistically meaningful difference among tokenizers in this setting. In the 2.5B setting, Unigram outperforms \tok by 0.29\%p on average. However, given the variance observed in the 560M experiments, this gap remains relatively small. These findings indicate that the proposed tokenizer maintains competitive English MCQA performance and does not substantially degrade English language understanding compared to other tokenizers.

\tok demonstrates stronger performance on the Korean MCQA benchmarks. In the 560M setting, \tok achieves the highest average score among the compared tokenizers, with an average of 53.09\%. Although the gap between \tok and BPE-Mecab is not statistically significant, \tok attains a comparable or slightly better average score. This outcome is notable because BPE-Mecab relies on an external Korean morphological analyzer, whereas \tok does not require language-specific preprocessing. The trend is more pronounced in the 2.5B setting, where \tok achieves the best average score of 57.43\%. It outperforms BPE and SuperBPE, indicating that the advantage observed in smaller-scale experiments persists at a larger model scale. These results demonstrate that \tok is particularly effective for Korean while maintaining a simpler and more language-agnostic tokenization pipeline.

\begin{table}[h]
\centering
\small
\resizebox{0.8\linewidth}{!}{
\begin{tabular}{@{}llccc@{}}
\toprule
\multirow{2}{*}{Lang.} & \multirow{2}{*}{Tokenizer}
& \multicolumn{2}{c}{IF-Eval} & GSM8K \\
\cmidrule(lr){3-4}
\cmidrule(lr){5-5}
& & Relaxed & Strict & EM (flexible) \\
\midrule
\multirow{4}{*}{\textbf{EN}}
& BPE       & 21.61 & 20.12 & 2.35 \\
& Unigram   & 23.42 & 22.33 & 1.90 \\
& SuperBPE  & \underline{25.95} & \textbf{25.83} & \textbf{2.50} \\
& \tok       & \textbf{27.99} & \underline{24.98} & \textbf{2.50} \\
\midrule
\multirow{5}{*}{\textbf{KO}}
& BPE       & 18.92 & 17.62 & 1.44 \\
& BPE-Mecab & 17.26 & 16.22 & 1.36 \\
& Unigram & \textbf{19.90} & \textbf{19.18} & \textbf{2.27} \\
& SuperBPE  & 16.59 & 16.59 & 1.21 \\
& \tok       & \underline{19.36} & \underline{18.35} & \underline{1.59} \\
\bottomrule
\end{tabular}
}
\caption{Evaluation results on English and Korean instruction-following and mathematical reasoning benchmarks. Korean results are evaluated on Ko-IF-Eval and Ko-GSM8K.}
\label{tab:generation_benchmarks}
\end{table}


Generation-oriented performance is further evaluated by instruction-fine-tuning the trained 2.5B models using publicly available synthetic datasets~\footnote{\url{https://huggingface.co/datasets/thunder-research-group/SNU_Thunder-synthetic-instruction-following}} based on KoAlpaca~\cite{alpaca} and SlimOrca~\cite{SlimOrca}. The datasets are generated using Exaone-3.5-32B-Instruct~\cite{exaone-3.5}, Llama-3.3-70B-Instruct~\cite{grattafiori2024llama}, and Qwen2.5-32B-Instruct~\cite{qwen2.5}. The results are presented in Table~\ref{tab:generation_benchmarks}.

For generation-oriented benchmarks, \tok demonstrates strong and consistent performance in both English and Korean. In English, \tok achieves the best IF-Eval relaxed score and ties for the best GSM8K score, while ranking second-best in the IF-Eval strict score, behind SuperBPE. In Korean, \tok delivers competitive results across all reported metrics, achieving the second-best performance on IF-Eval relaxed, IF-Eval strict, and Ko-GSM8K. Notably, \tok outperforms BPE, BPE-Mecab, and SuperBPE on all Korean generation-oriented benchmarks, indicating effectiveness beyond English-centric evaluation settings.

\subsection{Fertility and Inference Time}
\label{sec:inference}












\begin{table}[ht]
\centering
\scriptsize
\setlength{\tabcolsep}{3.5pt}
\begin{tabular}{llrrrrrr}
\toprule
\multirow{2}{*}{Lang.} & \multirow{2}{*}{Tokenizer}
& \multicolumn{3}{c}{Fertility $\downarrow$}
& \multirow{2}{*}{Time (s) $\downarrow$}
& \multirow{2}{*}{\#tok. $\downarrow$}
& \multirow{2}{*}{ms/tok} \\
\cmidrule(lr){3-5}
& & Train & Valid & Avg. & & & \\
\midrule
\multirow{4}{*}{\textbf{EN}}
& BPE      & 1.22 & 1.21 & 1.21 & 248.65 & 25,045 & 9.93 \\
& Unigram  & 1.39 & 1.39 & 1.39 & 283.15 & 28,649 & 9.88 \\
& SuperBPE & 0.97 & 0.95 & \underline{0.96} & 196.74 & 20,343 & 9.67 \\
& BTok     & 0.92 & 0.90 & \textbf{0.91} & 187.43 & 18,771 & 9.99 \\
\midrule
\multirow{5}{*}{\textbf{KO}}
& BPE       & 1.57 & 1.43 & 1.50 & 222.00 & 22,451 & 9.89 \\
& BPE-Mecab & 2.02 & 1.92 & 1.97 & 292.92 & 29,402 & 9.96 \\
& Unigram   & 1.95 & 1.84 & 1.89 & 277.30 & 28,310 & 9.80 \\
& SuperBPE  & 1.44 & 1.27 & \textbf{1.35} & 201.83 & 20,295 & 9.94 \\
& BTok      & 1.46 & 1.29 & \underline{1.37} & 204.15 & 20,572 & 9.92 \\
\bottomrule
\end{tabular}
\caption{
Fertility, inference time, and generated token counts across tokenizers.
Fertility is measured on 500 training sentences and 500 held-out validation sentences.
Inference time and token counts are measured on the same 1,000 target sentences.
}
\label{tab:generation_throughput}
\end{table}

\tok is designed to reduce fertility by representing frequent and structurally valid text spans with fewer tokens. The first step is to measure whether \tok produces more compact tokenizations than standard subword tokenizers on both the tokenizer-training corpus and a held-out validation corpus. The subsequent analysis examines whether this reduction in fertility results in a practical decrease in inference time. The inference time measurement serves as a validation step rather than as an independent assessment of model quality. Given the same model architecture and set of target texts, lower fertility should reduce the number of token-level generation steps required to produce an equivalent amount of text.

For each language, fertility is computed on 500 training sentences and 500 held-out validation sentences sampled from the corresponding language-specific XSum corpus~\cite{xsum-emnlp}. Fertility is defined as the number of tokens per whitespace-delimited unit; for Korean, this corresponds to an eojeol-level unit. The total number of generated tokens and the wall-clock inference time required to generate the same 1,000 target sentences are then measured. To distinguish the effect of sequence-length reduction from token-level inference throughput, the average inference time per token is also reported. The inference-time experiment utilizes the 560M model trained in Section~\ref{sec:performance}. All experiments are conducted on a single RTX~5090 GPU. The fertility and inference-time results for each tokenizer are summarized in Table~\ref{tab:generation_throughput}.

In English, \tok achieves approximately 25\% lower average fertility than BPE, reducing fertility from 1.21 to 0.91. This reduction is evident in the number of generated tokens: \tok requires 18,771 tokens to generate the same target sentences, whereas BPE requires 25,045 tokens. The corresponding inference time also decreases from 248.65 seconds to 187.43 seconds. Notably, the average inference time per token remains similar across tokenizers, indicating that the reduction in total inference time is primarily due to fewer generation steps rather than faster per-token computation.

In Korean, although SuperBPE yields slightly lower fertility than \tok, its language-modeling performance degrades substantially, as demonstrated in the previous section. This outcome indicates that reducing fertility alone is insufficient; the tokenizer must also preserve useful linguistic structure to maintain downstream model performance. In contrast, \tok reduces average fertility by approximately 9\% compared with BPE, from 1.50 to 1.37, while achieving language-modeling performance comparable to BPE-Mecab. The number of generated tokens decreases from 22,451 with BPE to 20,572 with \tok, and the inference time decreases from 222.00 seconds to 204.15 seconds. Notably, the fertility of \tok is 31\% lower than that of BPE-Mecab, despite BPE-Mecab's reliance on a Korean morphological analyzer.

Although \tok reduces fertility in both English and Korean, the greater reduction observed in English reflects language-specific differences in token distribution. A detailed analysis is provided in Appendix~\ref{app:token_distribution_analysis}.

\subsection{Ablation Study}
\label{sec:ablation}

\begin{table}[h]
\centering
\scriptsize
\setlength{\tabcolsep}{2.5pt}
\renewcommand{\arraystretch}{0.9}
\resizebox{\columnwidth}{!}{%
\begin{tabular}{@{}ccc cccc@{}}
\toprule
\multirow{2}{*}{\makecell{MW\\Cand.}}
& \multirow{2}{*}{\makecell{Seed\\Filter}}
& \multirow{2}{*}{\makecell{Token\\Score}}
& \multicolumn{2}{c}{EN}
& \multicolumn{2}{c}{KO} \\
\cmidrule(lr){4-5}
\cmidrule(lr){6-7}
& & & Perf.($\uparrow$) & Fert.($\downarrow$) & Perf.($\uparrow$) & Fert.($\downarrow$) \\
\midrule
\checkmark & \checkmark & \checkmark
& 41.56 {\scriptsize (0.39)} & 0.91
& 53.09 {\scriptsize (0.16)} & 1.37 \\

\checkmark & \xmark    & \checkmark
& 40.51 {\scriptsize (0.40)} & 0.96
& 50.50 {\scriptsize (0.23)} & 1.45 \\

\xmark     &  \xmark   & \checkmark
& 41.53 {\scriptsize (0.20)} & 1.23
& 52.43 {\scriptsize (0.41)} & 1.64 \\
\bottomrule
\end{tabular}%
}
\caption{Ablation study of the proposed tokenizer construction components.
MW(Multi-Word) Cand., Seed Filter, and Token Score denote the use of multi-word candidate tokens,
seed vocabulary filtering, and the proposed token scoring method, respectively.
Parenthesized values are standard deviations across repeated runs.}
\label{tab:ablation}
\end{table}

This section evaluates the effectiveness of the proposed seed vocabulary construction method and token scoring strategy. To assess the impact of removing incomplete tokens during seed vocabulary construction, an ablation is conducted that omits the seed filtering step. The effect of the proposed token scoring method is also evaluated by applying it under the same conditions as the Unigram tokenizer, replacing only the token scoring criterion. The results are presented in Table~\ref{tab:ablation}.

Omitting the seed vocabulary filtering step results in a substantial performance decrease in both languages. Compared with the full configuration, the model without seed filtering shows a performance reduction from 41.56\% to 40.51\% in English and from 53.09\% to 50.50\% in Korean, corresponding to decreases of 1.05 percentage points and 2.59 percentage points, respectively. These findings demonstrate that filtering out incomplete tokens is essential for constructing an effective seed vocabulary. Furthermore, although seed filtering imposes additional constraints on candidate tokens, it also reduces fertility from 0.96 to 0.91 in English and from 1.45 to 1.37 in Korean. This suggests that the proposed filtering strategy enhances both downstream model performance and tokenization efficiency.

Applying only the proposed token scoring method yields a tokenizer with performance statistically comparable to the full \tok configuration in both English and Korean. The tokenizer achieves 41.56\% in English and 52.43\% in Korean, closely matching the full model scores of 41.56\% and 53.09\%, respectively. These results indicate that the proposed token-scoring criterion is effective at selecting useful tokens. Notably, since BPE, which relies primarily on token frequency, underperforms \tok in Korean, the proposed scoring method appears to capture token utility more effectively than frequency alone. Compared with the standard Unigram tokenizer reported in Table~\ref{tab:generation_throughput}, the token-scoring-only variant does not produce a statistically significant performance difference but achieves substantially lower fertility. Specifically, fertility is reduced by 0.16 in English and 0.25 in Korean, while maintaining strong downstream performance. These findings demonstrate that the proposed token-scoring method reduces fertility without compromising language model performance.

\section{Conclusion}
This paper introduces \tok, a subword tokenizer that reduces token fertility without compromising language model performance. \tok constructs a large seed vocabulary from corpus substrings, filters structurally incomplete candidates at both the Unicode-byte and word-boundary levels, and prunes the vocabulary using a likelihood-based token score derived from a uniform Jensen lower bound. Experiments with 560M- and 2.5B-parameter language models demonstrate that \tok reduces fertility by approximately 25\% in English and 9\% in Korean compared with standard BPE, while maintaining competitive English performance and strong Korean performance without language-specific morphological preprocessing. These results indicate that compact and robust tokenization can be achieved by jointly considering structural validity and token utility.
\section*{Limitations}

This work has two main limitations. First, because our experiments train
language models from scratch, the computational cost limits the amount of
training, the number of downstream benchmarks, and the number of repeated
runs we can perform. While the results show consistent trends across
English and Korean, more extensive training and repeated evaluations
would provide a stronger estimate of the robustness of the observed
differences. Although we evaluate \tok on English and Korean, these two languages do not cover the full diversity of linguistic typology. Further evaluation on additional morphologically rich, low-resource, and scriptio continua languages is needed to establish broader cross-lingual generality

Second, applying \tok to pretrained language models is not
straightforward. Our preliminary experiments show that adapting only the
input embeddings and the language-modeling head is insufficient to obtain
the benefits of \tok, unlike what may be assumed in simple tokenizer
adaptation settings. This indicates that tokenizer replacement after
pretraining likely requires more substantial model adaptation, such as
continued pretraining or broader parameter updates. The results of these experiments are
reported in Appendix~\ref{app:tokenizer_adaptation}.
\section*{Ethics Considerations}

Our research focuses on improving language-model efficiency through tokenizer design. All datasets and benchmarks used in our experiments are publicly available or publicly distributable resources. For evaluation, we use standard English and Korean benchmarks such as ARC, HellaSwag, CommonsenseQA, PIQA, RACE, Ko-ARC, KoBEST HellaSwag, KoBEST COPA, and Ko-LAMBADA. For generation-oriented experiments, we use publicly available synthetic instruction-tuning data based on KoAlpaca and SlimOrca.

For language model training, we use data sources suitable for research and public release. In particular, because publicly available Korean training corpora are relatively limited, we construct the Korean training corpus by filtering Common Crawl data. This process does not involve additional human annotation or newly collected human-subject data. The filtered corpus is used solely for academic research, and we take care to avoid the use of private or sensitive information.

We follow the intended use and licensing conditions of all datasets and data sources used in this study. We do not attempt to infer private or sensitive information about individuals, and all experiments are conducted for academic research purposes.

\section*{Acknowledgments}
This work was partially supported by the National Research Foundation of Korea (NRF) under Grant No. RS-2023-00222663 (Center for Optimizing Hyperscale AI Models and Platforms) and under Grant No. A400-20260031, and by the Institute for Information and Communications Technology Promotion (IITP) under Grant No. 2018-0-00581 (CUDA Programming Environment for FPGA Clusters) and No. RS-2025-02304554 (Efficient and Scalable Framework for AI Heterogeneous Cluster Systems), all funded by the Ministry of Science and ICT (MSIT) of Korea. It was also partially supported by the Korea Health Industry Development Institute (KHIDI) under Grant No. RS-2025-25454559 (Frailty Risk Assessment and Intervention Leveraging Multimodal Intelligence for Networked Deployment in Community Care), funded by the Ministry of Health and Welfare (MOHW) of Korea. Additional support was provided by the BK21 Plus Program for Innovative Data Science Talent Education (Department of Data Science, Seoul National University, No. 5199990914569) and the BK21 FOUR Program for Intelligent Computing (Department of Computer Science and Engineering, Seoul National University, No. 4199990214639), both funded by the Ministry of Education (MOE) of Korea. This work was also partially supported by the Advanced GPU Utilization Support Program, funded by the Ministry of Science and ICT (MSIT) of Korea and operated by the National IT Industry Promotion Agency (NIPA). Research facilities were provided by the Institute of Computer Technology (ICT) at Seoul National University.



\bibliography{custom}

\clearpage
\appendix

\section{Seed Vocabulary Construction Details}
\label{app:initial_vocab_construction_details}
We describe the implementation details of the proposed filtering procedure used during seed vocabulary construction.
Given a set of preprocessed sentences with frequency counts, we first concatenate all sentences into a single flattened string. Each sentence is separated by a special sentence-boundary symbol \texttt{\textbackslash 0}, and an auxiliary frequency array is constructed so that each suffix-array position can be associated with the frequency of the sentence from which it originates. We then build a suffix array over the flattened string and enumerate candidate substrings from suffix-array intervals. This allows frequent substrings to be extracted efficiently from the corpus. Candidates that cross more than one sentence boundary are discarded, and candidates that encounter a sentence boundary are truncated at the first occurrence of \texttt{\textbackslash 0}.

We then apply Unicode decodability filtering. Since substring candidates are enumerated over the byte-level representation of the corpus, a candidate may start or end inside the UTF-8 byte sequence of a character. Our goal is not to remove every partial byte sequence, but to remove structurally mixed candidates that combine a complete Unicode string with an incomplete byte fragment of another character.

To implement this distinction, we first construct a mapping from each internal character representation to its underlying byte value. We then scan each candidate as a byte sequence and identify whether its bytes form complete UTF-8 characters or incomplete fragments of UTF-8 characters observed in the corpus. A candidate is retained if all of its bytes form a valid Unicode string. A candidate is also retained if it consists only of a contiguous byte
fragment from a single Unicode character, since such isolated partial-byte pieces may be needed to preserve byte-level coverage. However, a candidate is discarded if it contains both at least one complete Unicode character and an incomplete byte fragment of another character. For example, in the Korean word "서울", the candidate "<0x84><0x9C>" is retained because it is an isolated fragment of the character "서," whereas "<0x84><0x9C>울" is discarded because it combines an incomplete fragment of "서" with the complete character "울."" Candidates that cannot be interpreted either as a valid Unicode string or as an isolated fragment of a single UTF-8 character are also discarded.

After Unicode decodability filtering, we apply word-boundary integrity filtering. In the implementation, whitespace is represented by a dedicated space marker, \texttt{\textbackslash u\{0120\}}. If a candidate contains a space marker other than the initial one, it is truncated before the last space marker. This removes potentially incomplete trailing word fragments while preserving the longest prefix that remains aligned with valid word boundaries. Candidates that contain a space marker but do not start with it are discarded, ensuring that multi-word candidates do not begin inside a word.

Finally, we remove trivial or length-invalid candidates. Specifically, candidates of length one or less are discarded, and candidates that exceed the maximum allowed piece length are removed. The frequency score of each surviving candidate is computed by summing the sentence frequencies over the corresponding suffix-array interval.

After filtering, all single-character pieces observed in the corpus are added to the seed vocabulary to guarantee character coverage. The remaining substring candidates are sorted by decreasing frequency score, deduplicated, and inserted into the seed set until the predefined seed vocabulary size is reached.

\section{Token Scoring Derivation}
\label{app:token_scroing_derivation}

Given a training dataset $D$, we assume that $D$ consists of independent text chunks $Y$, such as sentences or short segments. Let $f(\cdot)$ denote empirical frequency. The log-likelihood of the dataset can then be written as

\begin{small}
\begin{equation*}
\log P(D) = \sum_{y \in D} f(y)\log P(y)
\end{equation*}
\end{small}

Let $X$ denote the tokenizer vocabulary. We interpret each chunk $y$ as being generated through the participation of vocabulary tokens $x \in X$. By applying the law of total probability, the likelihood of a chunk can be decomposed as

\begin{small}
\begin{align*}
\log P(y)
&=
\log \sum_{x \in X} P(y,x) \\
&=
\log |X|
+
\log \left(
\frac{1}{|X|}
\sum_{x \in X} P(y,x)
\right) \\
&\ge
\log |X|
+
\frac{1}{|X|}
\sum_{x \in X} \log P(y,x),
\end{align*}
\end{small}

where the inequality follows from Jensen’s inequality. Therefore, this gives a uniform Jensen lower bound on the log-likelihood, or equivalently an upper bound on the negative log-likelihood:

\begin{small}
\begin{align*}
\log P(D)
\ge
\sum_{y \in D} f(y)
\left(
\log |X| +
\frac{1}{|X|}
\sum_{x \in X} \log P(y,x)
\right)
\end{align*}
\end{small}

When a token $x$ is removed from the vocabulary, the only token-dependent component of this bound is

\begin{small}
\begin{equation*}
\sum_{y \in D} f(y)
\sum_{x \in X}
\log P(y,x)
\end{equation*}
\end{small}

Thus, within a pruning step, we use the decrease in this token-dependent term as a tractable proxy for the decrease in the uniform Jensen bound. This proxy ignores candidate-independent terms in the bound and relies on local updates to the statistics of the removed token and its replacement sequence. Therefore, the resulting score should be interpreted as an approximate greedy criterion rather than as the exact change in the full bound.

For each token $x$, let $Y_x$ denote the set of chunks in which $x$ appears. 
For analytical tractability, we further aggregate token occurrences at the chunk level and assume that each token appears at most once in each chunk. Under this approximation, the total frequency of chunks containing $x$ satisfies

\begin{small}
\begin{equation*}
\sum_{y \in Y_x} f(y) = f(x).
\end{equation*}
\end{small}

Substituting these definitions into the objective, we obtain

\begin{small}
\begin{align*}
&\sum_{y \in D} f(y)\log P(x, y) \\
&=
\sum_{x \in X}
\sum_{y \in Y_x}
f(y)
\left(
\log P(y \mid x) + \log P(x)
\right) \\
&=
\sum_{x \in X}
\left(
\sum_{y \in Y_x} f(y)\log P(y \mid x)
+
f(x)\log P(x)
\right)
\end{align*}
\end{small}

Since $P(y \mid x)$ represents the empirical probability that an
occurrence of $x$ comes from a text chunk $y \in Y_x$, we estimate it as

\begin{small}
\begin{equation*}
P(y \mid x) = \frac{f(y)}{f(x)},    
\end{equation*}
\end{small}

the objective can be simplified as

\begin{small}
\begin{align*}
\log P(D)
&\ge
\sum_{x \in X} f(x)\log P(x) \\
& +  \sum_{x \in X}
f(x)
\sum_{y \in Y_x}
P(y \mid x)\log P(y \mid x) \\
&=
\sum_{x \in X} f(x)\log P(x)
-
\sum_{x \in X} f(x)BE(x),
\end{align*}
\end{small}

where

\begin{small}
\begin{equation*}
BE(x) = -\sum_{y \in Y_x}
P(y \mid x)\log P(y \mid x)
\end{equation*}
\end{small}

denotes the branching entropy associated with token $x$.

Therefore, if we define $\Delta \ell(x)$ as the change in

\begin{small}
\begin{equation*}
\sum_{x \in X} f(x)\log P(x)
\end{equation*}
\end{small}

caused by removing token $x$, and define $\Delta BE(x)$ as the corresponding change in

\begin{small}
\begin{equation*}
\sum_{x \in X} f(x)BE(x),
\end{equation*}
\end{small}

then the Equation~\ref{eq:token_score_expanded} introduced in Method~\ref{sec:methods} follows directly.

\section{Detailed Computation of Token Scores}
\label{app:token_score_computation}

This section describes the local approximation used to compute the token scores in Equation~\ref{eq:token_score_expanded}. Rather than recomputing the full vocabulary statistics after removing each candidate token, we update only the statistics affected by replacing occurrences of $x$ with its retokenized sequence $Z_x$.

For a candidate token $x$, let $Z_x=(z_1,\ldots,z_m)$ denote the sequence of tokens obtained by
re-tokenizing the string represented by $x$ under the vocabulary
$X\setminus\{x\}$. The concatenation of $z_1,\ldots,z_m$ recovers the
original string represented by $x$.

\paragraph{Updated unigram probability.}
When $x$ is removed, each occurrence of $x$ is replaced by the token
sequence $Z_x$. This changes both the frequencies of the replacement
tokens and the total number of tokens. The updated unigram probability of
each replacement token $z \in Z_x$ is

\begin{small}
\begin{equation}
\label{eq:updated_unigram_probability}
P'(z)
=
\frac{f(z)+f(x)}
{\sum_{u \in X} f(u) + (|Z_x|-1)f(x)} .
\end{equation}
\end{small}

The denominator increases by $(|Z_x|-1)f(x)$ because each occurrence of
$x$ is replaced by $|Z_x|$ tokens.

\paragraph{Token-level log-likelihood change.}
Using the updated unigram probability, the token-level log-likelihood
change is computed as

\begin{small}
\begin{equation}
\label{eq:delta_token_likelihood}
\Delta \ell(x)
=
f(x)
\left(
\log P(x)
-
\sum_{z \in Z_x}\log P'(z)
\right).
\end{equation}
\end{small}

This term measures how much the unigram likelihood contribution decreases
when $x$ is represented by the replacement sequence $Z_x$.

\paragraph{Updated branching entropy.}
When $x$ is removed, its occurrences are redistributed to the replacement
tokens in $Z_x$. The updated frequency of each replacement token is

\begin{small}
\begin{equation}
\label{eq:updated_frequency}
f'(z)=f(z)+f(x).
\end{equation}
\end{small}

Let $P'(y\mid z)$ denote the updated conditional distribution over text
chunks $y$ given token $z$ after this redistribution. The updated
branching entropy is

\begin{small}
\begin{equation}
\label{eq:updated_branching_entropy}
BE'(z)
=
-\sum_{y \in Y} P'(y\mid z)\log P'(y\mid z).
\end{equation}
\end{small}

\paragraph{Branching-entropy change.}
The change in the branching-entropy contribution is

\begin{small}
\begin{equation}
\begin{split}
\label{eq:delta_branching_entropy}
\Delta BE(x)
& =
f(x)BE(x)
+
\sum_{z \in Z_x} f(z)BE(z) \\
& -
\sum_{z \in Z_x} f'(z)BE'(z).
\end{split}
\end{equation}
\end{small}

The first term is the original branching-entropy contribution of the
removed token $x$. The second term is the original contribution of the
replacement tokens. The final term is the updated contribution of the
replacement tokens after the occurrences of $x$ have been redistributed.

Finally, substituting Equations~\ref{eq:delta_token_likelihood} and
\ref{eq:delta_branching_entropy} into Equation~\ref{eq:token_score_expanded} gives the
token score used for vocabulary pruning.

\section{Tokenizer Configuration Details}
\label{app:tokenizer_details}

\begin{table*}[t]
\centering
\begin{tabular}{l|l}
\toprule
\textbf{Tokenizer} &\textbf{Segmented Text} \\
\midrule
\multicolumn{2}{l}{\textbf{English}} \\
\midrule
BPE        & t|-shirts| is| shown| at| an| outdoor| event| with| tents| at| night|. \\ 
Unigram    & t|-shirt|s| is| shown| at| an| outdoor| event| with| tents| at| night|.
 \\
SuperBPE  & t|-shirts| is| shown| at an| outdoor| event| with| tents| at night|. \\
{\tok}     & t-shirts| is shown| at| an outdoor| event| with| tents| at night. \\
\midrule
\multicolumn{2}{l}{\textbf{Korean}} \\
\midrule
BPE        & 타|투를| 받는| 동안|,| 아픈|지| 공을| 몇번| 세게| 쥔|다|. \\ 
BPE-Mecab  & 타투|를| 받|는| 동안|,| 아픈지| 공|을| 몇|번| 세|게| 쥔|다|. \\ 
Unigram    & 타투|를| 받는| 동안|,| 아픈|지| 공을| 몇번| 세게| 쥔|다|.
 \\
SuperBPE  & 타|투를| 받는| 동안|,| 아픈|지| 공을| 몇번| 세게| {\reuni}|{\reuni}다|. \\
{\tok}     & 타투|를| 받는| 동안,| 아픈|지| 공을| 몇번| 세게| 쥔|다. \\
\bottomrule
\end{tabular}
\caption{Examples of segmented text for each tokenizer. Tokens are delimited by vertical bars. The symbol \protect\reuni{} denotes an undecodable byte.}
\label{tab:tokenizer_comparsion}
\end{table*}

We apply a rule-based pre-tokenization scheme inspired by SuperBPE~\cite{liu2025superbpe}, which segments raw text into atomic units prior to subword learning using regular expressions.
This pre-tokenization is designed to avoid aggressive word-internal splitting while explicitly handling numeric sequences, non-alphanumeric symbol groups, and whitespace.

Specifically, the pre-tokenization rules:

\begin{itemize}
    \item split numeric sequences into chunks of up to three digits
    \item group consecutive non-letter and non-digit symbols into a single segment
    \item preserve trailing and isolated whitespace as independent units
\end{itemize}

This design encourages the tokenizer to learn longer and more semantically meaningful subword or multi-word tokens, while reducing unnecessary fragmentation caused by punctuation or formatting artifacts.

In addition, following the configuration that achieved the best performance in the original SuperBPE paper, the SuperBPE vocabulary is composed of 90\% tokens learned via standard BPE, with the remaining 10\% obtained by merging the tokens learned through the proposed method.

For Unigram and \tok, when constructing the seed vocabulary, we set the maximum length to 16 bytes for English and 48 bytes for Korean, considering that a single Korean character is typically encoded with 3 bytes. The number of initial candidates was set to 1,000,000.

\section{Model Architecture and Training Details}
\label{sec:experiment_details}

We use the LLaMA3~\cite{grattafiori2024llama} architecture with sequence length 1024 for all experiments. The architectural differences between the 560M and 2.5B models are summarized in Table~\ref{tab:model_config}.

\begin{table}[h]
\centering
\begin{tabular}{lcc}
\toprule
\textbf{Configuration} & \textbf{560M} & \textbf{2.5B} \\
\midrule
Hidden Size & 1024 & 2048 \\
Intermediate Size & 1024 & 8192 \\
Number of Attention Heads & 16 & 16 \\
Number of Hidden Layers & 24 & 32 \\
Training Steps & 40k & 80k \\
\bottomrule
\end{tabular}
\caption{Differences between the 560M and 2.5B LLaMA-based models used in our experiments}
\label{tab:model_config}
\end{table}

For all training, we use the AdamW optimizer~\cite{loshchilov2017decoupled} with $\beta_1 = 0.9$ and $\beta_2 = 0.999$. We apply a cosine learning rate scheduler~\cite{loshchilov2016sgdr} with 0.1\% of the total training steps reserved for warmup. The maximum learning rate is set to $5e-4$, and the minimum learning rate is $1e-8$. A batch size of 288 is used in all experiments. Model training is conducted using the DeepSpeed ZeRO~\cite{rajbhandari2020zero} data parallelism.

For the training corpora, we use FineWeb-Edu~\cite{lozhkov2024fineweb-edu} for English and a Korean CommonCrawl corpus filtered following the procedure of \citet{park2025beyond}. For tokenizer training, we randomly sample 100K documents from each language corpus. For model training, we train the 560M models on 10B tokens and the 2.5B models on 50B tokens.

We use token-budget matching rather than raw-epoch matching. Let $C$ denote the shared raw training corpus and let $T_i(C)$ denote the token sequence obtained by applying tokenizer $i$ to $C$. Since $|T_i(C)|$ differs across tokenizers, especially for low-fertility
tokenizers, we construct each training stream by repeatedly cycling through $T_i(C)$ until the desired token budget is reached. The training budget, number of optimization steps, batch size, and model architecture are kept fixed across tokenizers. Therefore, all models
are trained on the same unique raw corpus, but compact tokenizers may cycle through the corpus slightly more often because they produce fewer tokens per raw sentence.

For the 560M model, we repeat each experiment five times and report the aggregated results, whereas for the remaining model settings, we report the result from a single run. The 560M models are trained on NVIDIA RTX 4090 and 5090 nodes, while the remaining models are trained on NVIDIA H100 and B200 nodes.

\section{Description of Benchmarks}
\label{sec:benchmarks}

Information on the sixteen benchmarks used in our experiments is summarized in Table~\ref{tab:benchmark_summary}.
All experiments were conducted using lm-eval-harness~\cite{eval-harness} version 0.4.13.
For benchmarks whose evaluation scores are highly sensitive to the length of answer choices, we adopted Normalized Probability Shift by Question (NPSQ), as proposed by \citet{cho2025choices}.

For CommonsenseQA, we convert the task into a cloze-style format, where the question is provided as input and the model is asked to generate the answer choice. For the Korean benchmarks released by \citet{mcrlkorean2025}, we use the same prompt templates as their corresponding English benchmarks whenever applicable, so that the English and Korean evaluations differ primarily in language rather than in evaluation format.

For PIQA and CLIcK, no explicit license information was available at the time of our review, and we therefore mark the license field as "--" in Table~\ref{tab:benchmark_summary}. 
We used these datasets only for model evaluation and did not use them for training or redistribution.

\begin{table*}[ht]
\centering
\small
\begin{tabularx}{\textwidth}{lllcccl}
\toprule
\textbf{Lang.} & \textbf{Benchmark} & \textbf{Task Type} & \textbf{\# Questions} & \textbf{Metric} & \textbf{\# Fewshot} & \textbf{License} \\ \midrule
\multirow{8}{*}{\textbf{EN}} & ARC-Challenge & Science QA & 2,376 & acc\_npsq & 25 & CC BY-SA 4.0 \\
 & HellaSwag & Commonsense Reasoning & 10,042 & acc\_npsq & 0 &MIT \\
 & Winogrande & Contextual Reasoning & 1,767 & acc & 5 & Apache 2.0 \\
 & CommonsenseQA & Commonsense Reasoning & 1,221 & acc & 0 & MIT \\
 & PiQA & Physical Commonsense & 1,838 & acc\_npsq & 0 & - \\
 & RACE (High) & Reading Comprehension & 3,498 & acc & 0 & Apache 2.0 \\ 
 & IF-Eval & Instruction Following & 541 & loose,strict & 0 & Apache 2.0 \\
 & GSM8k & Math Problem Solving & 1,319 & exact match & 5 & MIT \\
 \midrule
\multirow{8}{*}{\textbf{KO}} & Ko-ARC-Challenge & Science QA & 2,376 & acc\_npsq & 5 & CC BY-SA 4.0 \\
 & Kobest-HellaSwag & Commonsense Reasoning & 500 & acc\_npsq & 0 & CC BY-SA 4.0 \\
 & Ko-Winogrande & Contextual Reasoning & 1,267 & acc & 0 & Apache 2.0 \\
 & CLIcK & Korean Culture and Linguistic QA & 1,995 & acc\_npsq & 0 & - \\ 
 & Kobest-COPA & Causal Reasoning & 1,000 & acc & 0 & CC BY-SA 4.0 \\ 
 & Ko-LAMBADA & Language Modeling & 2,260 & acc & 0 & CC BY-SA 4.0 \\
 & Ko-IF-Eval & Instruction Following & 841 & loose,strict & 0 & Apache 2.0 \\
 & Ko-GSM8k & Math Problem Solving & 1,319 & exact match & 5 & MIT \\
 \bottomrule
\end{tabularx}
\caption{Summary of English and Korean Benchmarks}
\label{tab:benchmark_summary}
\vspace{1mm}
\end{table*}

\section{Token Distribution Analysis}
\label{app:token_distribution_analysis}

\begin{figure}[t]
    \centering
    \includegraphics[width=\linewidth]{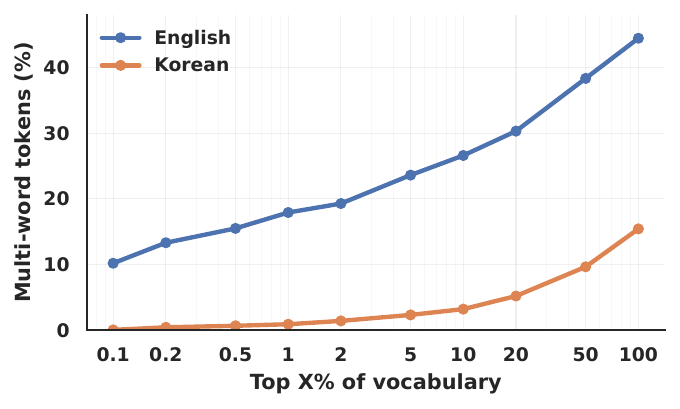}
    \caption{
    Cumulative proportion of multi-word tokens among the top-ranked vocabulary entries of \tok. 
    }
    \label{fig:multiword_ratio}
\end{figure}

We further analyze why the fertility reduction of \tok is larger in English than in Korean. Fertility decreases when frequent units in the corpus can be represented as single vocabulary items. In English, many frequent patterns correspond to multi-word expressions or short function-word sequences, such as "of the" and "on the." Since conventional tokenizers typically construct tokens within word boundaries, they cannot directly represent these frequent cross-word patterns as single tokens. By allowing tokens to span multiple words, \tok can include such expressions in the vocabulary, resulting in a larger fertility reduction.

Figure~\ref{fig:multiword_ratio} supports this interpretation. English contains a substantially higher proportion of multi-word tokens among the top-ranked vocabulary entries of \tok than Korean. In particular, multi-word tokens occupy a larger fraction of the high-ranked region of the English vocabulary, indicating that many frequent English expressions can be compressed into single tokens. This suggests that cross-word tokenization provides more opportunities for reducing fertility in English.

In Korean, frequent grammatical elements such as particles and endings are attached to content words without whitespace. As a result, frequent patterns are more likely to occur within eojeol-level units rather than across whitespace-separated words, leaving fewer opportunities for fertility reduction through multi-word tokens alone. This helps explain why \tok achieves a smaller fertility reduction in Korean than in English. Although SuperBPE achieves slightly lower Korean fertility, its downstream performance is weaker, suggesting that more aggressive compression does not necessarily preserve token units that are useful for language modeling.

\section{Tokenizer Adaptation}
\label{app:tokenizer_adaptation}

\begin{table*}[t]
\centering
\small
\renewcommand{\arraystretch}{0.9}
\begin{tabular}{@{}l@{\hspace{2pt}}l@{\hspace{5pt}}cccccccc@{}}
\toprule
\multirow{3}{*}{\textbf{Lang.}} 
& \multirow{3}{*}{\textbf{Tokenizer}}
& \multirow{3}{*}{\makecell[c]{\textbf{Tokenizer}\\\textbf{Overlap}}}
& \multicolumn{7}{c}{\textbf{MCQA Avg.}} \\
\cmidrule(l){4-10}
& 
& 
& \textbf{Emb.+LM Head}
& \multicolumn{6}{c}{\textbf{Full Model}} \\
\cmidrule(l){4-10}
&
&
& \textbf{2B}
& \textbf{2B}
& \textbf{10B}
& \textbf{20B}
& \textbf{30B}
& \textbf{40B}
& \textbf{50B} \\
\midrule
\multirow{5}{*}{EN}
& Qwen     & --      & \multicolumn{7}{c}{59.03} \\ \cmidrule(l){2-10} 
& BPE      & 40.78\% &  58.84  &  56.83 & 57.30 & 56.79 & 57.74 & 57.34 & 59.12 \\
& Unigram  & 27.74\% &  56.90  &  56.26 & 56.92 & 57.19 & 57.50 & 57.47 & 57.77 \\
& SuperBPE & 39.40\% &  56.42  &  55.43 & 55.14 & 55.47 & 55.61 & 55.54 & 56.88 \\
& \tok     & 28.15\% &  53.67  &  54.83 & 55.41 & 55.63 & 55.80 & 56.01 & 57.08 \\
\midrule
\multirow{6}{*}{KO}
& Qwen      & --      & \multicolumn{7}{c}{50.99} \\ \cmidrule(l){2-10}
& BPE       & 7.22\%  &  51.03  &  57.61 & 58.93 & 59.58 & 60.19 & 60.76 & 60.78 \\
& Unigram   & 6.76\%  &  52.58  &  57.68 & 60.01 & 60.22 & 60.75 & 61.36 & 61.63 \\
& BPE-Mecab & 13.18\% &  52.60  &  55.38 & 57.30 & 58.54 & 58.88 & 59.15 & 58.97 \\
& SuperBPE  & 6.40\%  &  50.71  &  56.49 & 58.78 & 59.49 & 59.89 & 60.03 & 59.95 \\
& \tok      & 5.81\%  &  51.43  &  57.23 & 59.26 & 60.12 & 60.32 & 60.61 & 60.76 \\
\bottomrule
\end{tabular}
\caption{Performance comparison of tokenizer adaptation on English and Korean MCQA benchmarks, along with token overlap ratios with the Qwen2.5-1.5B tokenizer. Qwen denotes the original model performance before tokenizer replacement. Emb.+LM Head denotes adaptation where only the input embedding and LM head layers are trained, while Full Model denotes continued training of all model parameters for 50B tokens. MCQA Avg. denotes the average score over multiple-choice QA benchmarks.}
\label{tab:tokenizer_adaptation}
\end{table*}

Previous tokenizer adaptation studies~\cite{li2025tokalign, gu2024retok} have shown that, after replacing the tokenizer, training only the embedding and LM head layers for up to 2B tokens is often sufficient to recover the performance of the original model. We examine whether this adaptation strategy is also effective for the tokenizers considered in this work, including \tok. 

We conduct tokenizer adaptation experiments using Qwen2.5-1.5B. For each new tokenizer, we replace the input embedding and LM head layers so that they match the corresponding vocabulary. If a token is shared with the original Qwen2.5 tokenizer, we initialize its embedding and LM head vectors by copying the original vectors. For newly introduced tokens, we tokenize the corresponding string using the original Qwen2.5 tokenizer and initialize its vectors by averaging the vectors of the resulting original tokens.

We evaluate two adaptation settings. In embedding-only adaptation, only the input embedding and LM head layers are trained for 2B tokens. In full-model continued training, all model parameters are updated for 50B tokens. In both settings, we use a learning rate of $2 \times 10^{-5}$, while all other training hyperparameters follow Appendix~E. The results are reported in Table~\ref{tab:tokenizer_adaptation}.

In contrast to prior findings, we observe that for English, where Qwen2.5 already exhibits strong performance, training only the embedding and LM head layers is not sufficient to recover the original model performance when the tokenizer is substantially changed. This is particularly evident for SuperBPE and \tok, both of which introduce multi-word tokens and therefore differ more significantly from the original tokenizer. However, when we continue training the entire model for approximately 50B tokens, the performance is partially recovered on MCQA benchmarks.

For Korean, on the other hand, we find that the original performance can be recovered even when only a subset of the model parameters is trained. This may be because Korean is not one of the primary languages on which Qwen2.5 was trained, leaving more room for adaptation through limited parameter updates.

These results suggest that the effectiveness of existing tokenizer adaptation methods may depend strongly on the degree of mismatch between the original and new tokenizers, as well as on the language being adapted. Prior approaches appear to work reasonably well when the new tokenizer remains close to the original BPE tokenizer, or when adaptation is performed for languages that are relatively underrepresented in the pretrained model. In contrast, when the vocabulary structure changes substantially, as in SuperBPE and \tok, successful tokenizer adaptation may require more extensive model updates. Developing adaptation methods that remain effective under such substantial tokenizer changes is an important direction for future work.

\end{document}